\documentclass[journal]{new-aiaa}
\usepackage[utf8]{inputenc}
\usepackage{textcomp}

\usepackage[usenames,dvipsnames]{color} 

\usepackage[capitalize]{cleveref}
\usepackage{longtable}
\usepackage{multicol}
\usepackage{pdfpages}
\usepackage[utf8]{inputenc}
\usepackage[table]{xcolor}
\usepackage[final]{changes}

\usepackage{graphics}
\usepackage{subfigure}
\usepackage{tikz}

\usepackage{pgfplots}
\usetikzlibrary{pgfplots.fillbetween}
\pgfplotsset{scaled x ticks=false}
\pgfplotsset{compat=newest}
\pgfplotsset{every axis/.append style={axis on top}}
\usepgfplotslibrary{groupplots}

\usepackage{nomencl}
\usepackage{cancel}

\usepackage[none]{hyphenat}

\usepackage[style=ieee, url=false, doi=true]{biblatex}

\usepackage{siunitx}
\usepackage{longtable,tabularx}
\usepackage{todonotes}
\title{Inferring Traffic Models in Terminal Airspace \\ from Flight Tracks and Procedures}

\author{Soyeon Jung,\footnote{Ph.D. candidate, Aeronautics and Astronautics.} Amelia Hardy,\footnote{M.S. student, Computer Science.} and Mykel J. Kochnederfer\footnote{Associate Professor, Aeronautics and Astronautics. AIAA Associate Fellow.  Corresponding author: mykel@stanford.edu.}}
\affil{Stanford University, Stanford, CA 94304 USA}

\begin{document}

\maketitle
\vspace{-1.2cm}
\begin{center}
    DOI: \href{https://doi.org/10.48550/arXiv.2303.09981}{10.48550/arXiv.2303.09981}
\end{center}

\begin{abstract}
Realistic aircraft trajectory models are useful in the design and validation of air traffic management (ATM) systems. 
Models of aircraft operated under instrument flight rules (IFR) require capturing the variability inherent in how aircraft follow standard flight procedures. The variability in aircraft behavior differs among flight stages. In this paper, we propose a simple probabilistic model that can learn this variability from procedural data and flight tracks collected from radar surveillance data. For each segment, we use a Gaussian mixture model to learn the deviations of aircraft trajectories from their procedures. Given new procedures, we generate synthetic trajectories by sampling a series of deviations from the Gaussian mixture model and reconstructing the aircraft trajectory using the deviations and the procedures. We extend this method to capture pairwise correlations between aircraft and show how a pairwise model can be used to generate traffic involving an arbitrary number of aircraft. We demonstrate the proposed models on the arrival tracks and procedures of the John F. Kennedy International Airport. Distributional similarity between the original and the synthetic trajectory dataset was evaluated using the Jensen-Shannon divergence between the empirical distributions of different variables and we provide qualitative analyses of the synthetic trajectories generated. 
\end{abstract}

\newlength\figureheight
\newlength\figurewidth

\section{Introduction}
\lettrine{D}{eveloping} and assessing air traffic management (ATM) systems is a key step towards increasing the safety of air transport that requires probabilistically modeling future aircraft trajectories based on past data.
The Federal Aviation Administration's (FAA) Next Generation Air Transportation System (NextGen) \cite{FAAnextgen} and EUROCONTROL's Single European Sky ATM Research Program (SESAR) \cite{Sesar}, which have led the modernization of global air traffic management (ATM) systems, are examples of such model-dependent ATM programs.
The concept of trajectory-based operations (TBO) is a cornerstone of these systems, managing air traffic based on four-dimensional trajectory (4DT) information which consists of a series of three-dimensional coordinates (latitude, longitude, and altitude) with an added time variable. However, open-source, real-world flight data is limited or inaccessible. This necessitates
constructing robust aircraft trajectory models, which are important for the successful implementation of modern ATM systems. 
These trajectory models can be useful for developing new concepts of operation, flight planning, aircraft scheduling, and conflict detection. All of these tasks are critical for the safety, efficiency, and predictability of air transport and are dependent on effective trajectory models.

Trajectory models should be able to represent the behavior of aircraft following real operations.
Aircraft operated under instrument flight rules (IFR) are instructed to follow standard flight procedures, defined as a series of virtual waypoints.
However, the actual paths flown by each aircraft vary due to various factors such as pilot behavior, weather, and preferences of air traffic controllers. 
Aircraft trajectory models should capture both this variability and the general tendency of aircraft to follow procedures.

An additional factor these models must account for is that the variability of aircraft behavior differs between flight stages.
For example, during the initial approach, aircraft tend to have high variability in following procedures because they are converging from multiple directions to the landing runway, and they can be radar vectored to the final approach course. By contrast, aircraft on the final approach have low variability because they need to stay aligned with the runway to land safely.

Various approaches have been proposed for modeling aircraft trajectories.
Traditional methods explicitly model aircraft behavior based on kinematic equations of motion.
Several studies predicted the nominal trajectory by propagating the state estimates into the future \cite{chatterji1996route, chatterji1999short} or to the next flight segments \cite{slattery1997trajectory}.
Aircraft performance models, such as Base of Aircraft Data (BADA) \cite{nuic2010user}, were also developed based on aircraft dynamics models for trajectory simulation and prediction.
These physics-based approaches do not account for the uncertainties that lie in aircraft trajectory prediction \cite{yepes2007new}. Furthermore, they are unable to capture the multimodality that is seen in real-world data.

Other efforts have developed probabilistic methods for modeling aircraft behavior.
Some studies involved learning aircraft dynamics or navigational intent based on dynamic Bayesian networks to account for the uncertainty in the future states of aircraft \cite{liu2011probabilistic,kochenderfer2008uncorrelated,weinert2013uncorrelated,  lowe2015learning, mahboubi2017learning}.
Researchers also applied supervised learning algorithms such as spline functions \cite{kun20084}, neural networks \cite{le1999using, hamed2013statistical}, and generalized linear models (GLM) \cite{de2013machine} to predict future trajectories or the estimated times of arrival. Although these models consider uncertainty, they do not capture how such uncertainty varies depending on flight stage.

The previously applied approaches offer interpretability and efficiency; however, they present relatively simple models that do not take advantage of the full amount of historical data available. Work that addresses these limitations includes recent studies that present recurrent neural networks (RNNs) to learn spatiotemporal patterns of aircraft trajectories \cite{liu2018predicting, pang2019aircraft}.
Another study applied generative adversarial imitation learning (GAIL) to learn the optimal policy given historical trajectories from expert demonstrations \cite{bastas2020data}.
Other researchers proposed unsupervised clustering algorithms such as $k$-means clustering \cite{gariel2011trajectory}, hierarchical clustering \cite{hong2015trajectory}, and Gaussian mixture models (GMMs) \cite{mahboubi2017learning, Barratt2019}. 
Some of them clustered turning points extracted from trajectories represented as a sequence of these clusters \cite{gariel2011trajectory} or transitions from one another \cite{mahboubi2017learning}.
Others clustered the entire trajectories directly using position measurements \cite{hong2015trajectory, conde2016trajectory, Barratt2019}. An advantage of such clustering-based approaches is that they can be used to generate synthetic trajectories based on historical data. However, clustering approaches define normative behavior as that which is most similar to previous observation and thus do not account for situations when deviations are more or less appropriate.

Recent work has leveraged long short-term memory (LSTM) networks for the trajectory prediction task 
 \cite{schimpf2023generalized, shi20204, yang2023aircraft}. To address instances when Automatic Dependent Surveillance Broadcast (ADS-B) technology is unavailable, such as in the event of an onboard equipment failure, researchers train a bidirectional LSTM on historical ADS-B data \cite{yang2023aircraft}. Flights may also deviate from expected patterns due to inclement weather, a challenge that researchers have approached by including weather data in their LSTM model \cite{schimpf2023generalized}. Others segment trajectories into climbing, cruising, and descending/approaching phases, training a unique LSTM network for each phase to allow optimizing for that phase's distinct characteristics \cite{shi20204}. Like the method we propose, such LSTM-based approaches can be used to produce synthetic trajectories. In our work, we also segment trajectories; however, the stages selected are different, as described in Section \ref{sec:dataset}. 

 Encoder-decoder-based architectures, such as variational autoencoders (VAE) and transformers have also been applied to trajectory prediction \cite{dong2023tcn, krauth2023deep, guo2022flightbert, pang2022bayesian}. When applying VAE networks to the trajectory prediction task, some researchers focus on the Terminal Manoeuvre area near Zurich airport, where aircraft are frequently observed deviating from nominal approach procedures \cite{krauth2023deep}. Their work shows these models to be highly effective in handling this uncertainty. To address trajectory prediction for multiple aircraft, other researchers use a Bayesian spatiotemporal graph transformer model \cite{pang2022bayesian}. This model combines a spatial transformer with a temporal transformer, adding Bayesian linear layers to the decoder to model trajectory uncertainty. Another hybrid approach combines a temporal convolutional network (TCN) with a transformer-based Informer model \cite{dong2023tcn}. Rather than formulating trajectory prediction as a time sequence prediction problem, other researchers define the task as a Multi Binary Classification (MBC) problem by transforming input and output trajectories into Binary Encoding (BE) representations \cite{guo2022flightbert}. Each scalar value of a flight trajectory attribute is converted into this format. Then, a Transformer embeds these binary representations and a predictor network makes a series of binary classifications to produce the output trajectory. These large models are powerful; however, such approaches are also highly complex and require tuning many hyper-parameters. 
 
 Complex, neural network models such as the RNN, LSTM, VAE, and Transformer-based models discussed all demonstrate promising results, but introduce potential for overfitting and decrease the potential for interpretability. Such complex models are best used when a simpler model is unable to capture the complexity seen in the real-world data modeled. In our work, we propose leveraging known flight procedures as the basis for a simpler, GMM-based model. We find that despite its lesser complexity, this model remains effective in modeling the flight trajectory data studied. 
 
Much of the prior work on trajectory models has not focused on the relationship between aircraft trajectories and flight procedures. The procedures, however, are the basis of air traffic control that facilitates control over many aircraft.
In our recent work \cite{Jung2019}, we proposed a probabilistic model that learns aircraft behavior in relation to procedures directly from recorded radar flight tracks and standard procedural data. 
We fit a GMM to learn a sequence of deviations between aircraft trajectory points and corresponding points on the flight procedure.
To accommodate the varying relationship between aircraft trajectories and procedures over multiple flight stages, we segmented aircraft trajectories for each flight stage and fit a separate model for each segment. This model is simple, trains quickly, and does not require hyperparameter tuning. 

This paper extends the trajectory model from our prior work to incorporate multiple aircraft into full traffic scenes. Full traffic scenes are especially important when validating the safety and operational efficiency of new air traffic control operations and concepts. Developing models that explicitly address these cases is crucial, as aircraft within close proximity influence each other's behavior. Thus, repeatedly applying a single trajectory model is not sufficient. 
We fit GMMs for pairs of trajectories and generate multiple trajectories by combining the mean vectors and covariance matrices of the pairwise GMMs.
This approach allows us to efficiently generate an arbitrary number of trajectories using a single model. Being able to predict and simulate these trajectories has many practical real-world applications. For example, a model like the one our work proposes can be used to forecast trajectories in advance or to support scheduling decisions by predicting landing time. To help avoid feature redundancy within and improve the robustness of our model, we further develop the trajectory model by performing a low-rank approximation for the covariance matrices in GMMs.

The remainder of this paper is structured as follows. Section \ref{sec:dataset} describes the trajectory and procedure data.
Section \ref{sec:single trajectory model} outlines the single trajectory model with discussions on the approximation method and synthetic trajectory generation method.
Section \ref{sec:multiple trajectory model} then extends the single trajectory model to incorporate multiple aircraft.
Section \ref{sec:experiments} presents experimental results, and section \ref{sec:conclusions} closes with a summary and future works. Our main contributions are as follows:
\begin{itemize}
    \item For a single trajectory scenario, a separate Gaussian mixture model is trained for each flight stage, to model the differences in variance across stages.
    \item We extend these single trajectory GMMs to pairwise GMMs, which we use to model trajectories of multiple aircraft in proximity. Then, we show how this model can be used to generate multiple trajectories within a single scenario.
    \item We train our proposed models using KJFK arrival trajectories. We evaluate these models on a test split of this data. Additionally, we study how well these models generalize by evaluating them on Charlotte Douglas Airport (KCLT) data. 
    \item Finally, we present both quantitative and qualitative analyses of our models' performance. Our results suggest that the proposed models are effective \textcolor{blue}{in }predicting both major patterns and uncertainties.
\end{itemize}

\section{Trajectory and Procedure Data}
\label{sec:dataset}
\subsection{Trajectory Data}
In this paper, we use a trajectory dataset that is generated using the Federal Aviation Administration (FAA) multisensor fusion tracker \cite{jagodnik2008fusion}.
This tracker collects radar detections from multiple sensors (primary radar, mode A/C and S transponders, wide area multilateration systems, and ADS-B receivers) and fuses them into a set of flight tracks.
Each track contains timestamped entries of the target address (used to group entries into flights), aircraft ID (if available), latitude and longitude (in WGS84), pressure altitude, geometric altitude (if available), and horizontal and vertical velocities.
The data were collected for six months starting March 2012 from three locations: Central Florida, New York City, and Southern California.
We train our models using data from 
the vicinity of John F. Kennedy International Airport (KJFK), located within the New York Class B airspace that includes Newark Liberty International Airport (KEWR) and LaGuardia Airport (KLGA).

\begin{figure}[bt]
\centering
\setlength\figureheight{7cm}
\setlength\figurewidth{7cm}
\begin{tikzpicture}

\begin{axis}[
colorbar,
colorbar style={width=7,
ytick={0,1,2,3,4},
yticklabel={\normalsize $10^{\pgfmathprintnumber{\tick}}$}},
colormap = {whiteblack}{color(0cm)=(white); color(1cm)=(black)},
height=\figureheight,
point meta max=4.5,
point meta min=0,
tick align=outside,
tick pos=left,
width=\figurewidth,
x grid style={black},
xlabel={East (NM)},
xmin=-20, xmax=25,
y grid style={black},
ylabel={North (NM)},
ymin=-25, ymax=20,
tick label style={font=\tiny},
label style={font=\scriptsize}
]

\addplot graphics [includegraphics cmd=\pgfimage,xmin=-30, xmax=30, ymin=-30, ymax=30] {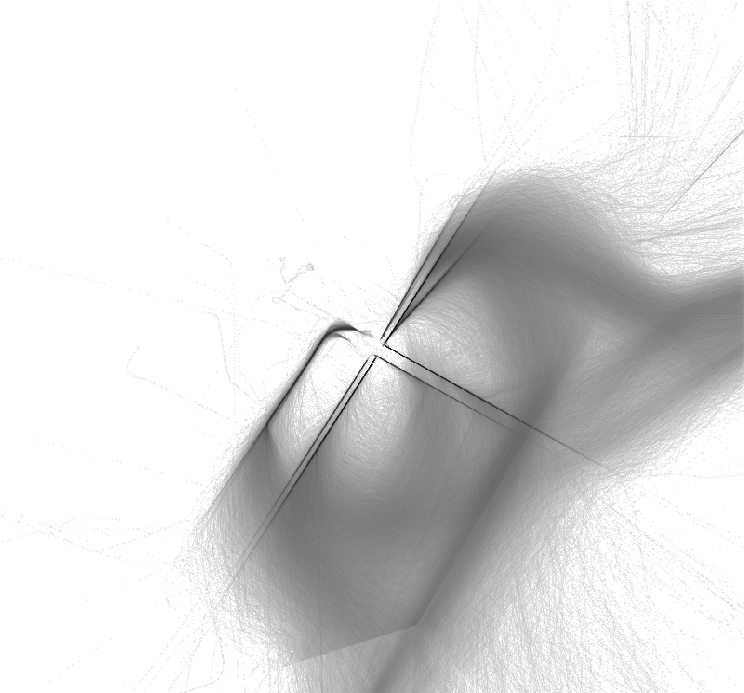};

\addplot [draw=none, draw=black, fill=black, colormap/viridis]
table [row sep=\\]{%
x                      y\\ 
+0.000000000000000e+00 -2.500000000000000e-01\\ 
+6.630077500000001e-02 -2.500000000000000e-01\\ 
+1.298949676962134e-01 -2.236584228970604e-01\\ 
+1.767766952966369e-01 -1.767766952966369e-01\\ 
+2.236584228970604e-01 -1.298949676962134e-01\\ 
+2.500000000000000e-01 -6.630077500000001e-02\\ 
+2.500000000000000e-01 +0.000000000000000e+00\\ 
+2.500000000000000e-01 +6.630077500000001e-02\\ 
+2.236584228970604e-01 +1.298949676962134e-01\\ 
+1.767766952966369e-01 +1.767766952966369e-01\\ 
+1.298949676962134e-01 +2.236584228970604e-01\\ 
+6.630077500000001e-02 +2.500000000000000e-01\\ 
+0.000000000000000e+00 +2.500000000000000e-01\\ 
-6.630077500000001e-02 +2.500000000000000e-01\\ 
-1.298949676962134e-01 +2.236584228970604e-01\\ 
-1.767766952966369e-01 +1.767766952966369e-01\\ 
-2.236584228970604e-01 +1.298949676962134e-01\\ 
-2.500000000000000e-01 +6.630077500000001e-02\\ 
-2.500000000000000e-01 +0.000000000000000e+00\\ 
-2.500000000000000e-01 -6.630077500000001e-02\\ 
-2.236584228970604e-01 -1.298949676962134e-01\\ 
-1.767766952966369e-01 -1.767766952966369e-01\\ 
-1.298949676962134e-01 -2.236584228970604e-01\\ 
-6.630077500000001e-02 -2.500000000000000e-01\\ 
+0.000000000000000e+00 -2.500000000000000e-01\\ 
+0.000000000000000e+00 -2.500000000000000e-01\\
};

\addplot [blue, line width=0.75pt, forget plot]
table [row sep=\\]{%
-0.314790350956774	-1.07388212440464 \\
0.701926179643114	0.634446106243 \\
};
\addplot [blue, line width=0.75pt, forget plot]
table [row sep=\\]{%
0.381018579296101	-0.869410987002108 \\
1.0883680175494	0.318538875436125 \\
};
\addplot [blue, line width=0.75pt, forget plot]
table [row sep=\\]{%
-0.527871725321972	1.06913560548559 \\
0.886517107609026	0.227348929251132 \\
};
\addplot [blue, line width=0.75pt, forget plot]
table [row sep=\\]{%
-1.73637529142706	0.505844290966593 \\
0.315676001958449	-0.715917614289667 \\
};
\end{axis}

\end{tikzpicture}
\caption{Log-histogram of all actual arrival tracks of KJFK.}
\label{fig:jfk_arrivals_hist}
\end{figure}
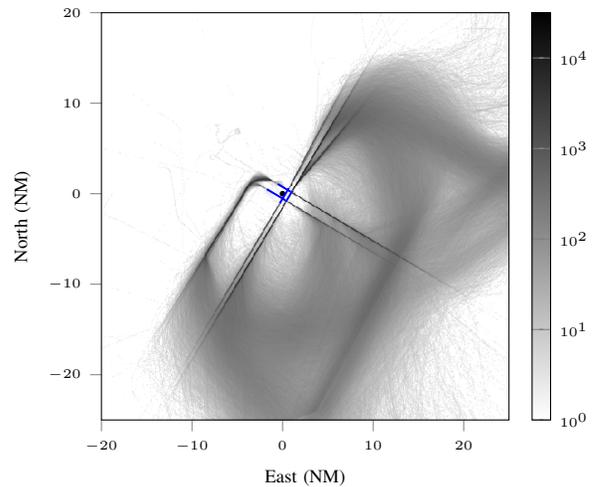

We process the track data as follows. First, the latitude-longitude-altitude coordinates of each position measurement are transformed to the local east-north-up (ENU) coordinates centered at the airport.
Next, we extract trajectories that enter into the airspace defined as 25 NM from the airport center. Based on the change in distance to the airport and altitude over time, the trajectories are sorted into arrivals, departures, and overflights
Finally, we separate the trajectories based on the runways: 04L/22R, 04R/22L, 13L/31R, and 13R/31L.
Fig. \ref{fig:jfk_arrivals_hist} shows the lateral view of a log-histogram of all arrival flight tracks to KJFK.
The origin indicates the center of the airport and the blue lines indicate the runways in use.

\subsection{Flight Procedure Data}
To model flight procedures, we use the standard flight procedures published by the air navigation service providers (e.g., the FAA). 
The procedures are defined as a set of waypoints, which are fixed points in 2D space (latitude and longitude). Each aircraft is instructed to fly from one waypoint to the next along the procedure.
Arrival traffic involves two types of standard flight procedures: standard terminal arrival routes (STARs) and instrument approach procedures (IAPs).
A STAR connects the end of an airway from the en-route airspace to the vicinity of the airport.
An IAP is a sequence of predetermined maneuvers used for the final approach. It allows the pilots to align the aircraft with the runway, make their final descent, and land safely in low visibility conditions. 
For the integration of multiple traffic approaches and the smooth transition of each aircraft from a STAR to an IAP, air traffic controllers (ATC) often provide radar vectoring (i.e., guide aircraft by assigning headings, altitudes, and speeds).

\begin{figure}[bt]
\centering
\setlength\figureheight{7cm}
\setlength\figurewidth{7cm}
\input{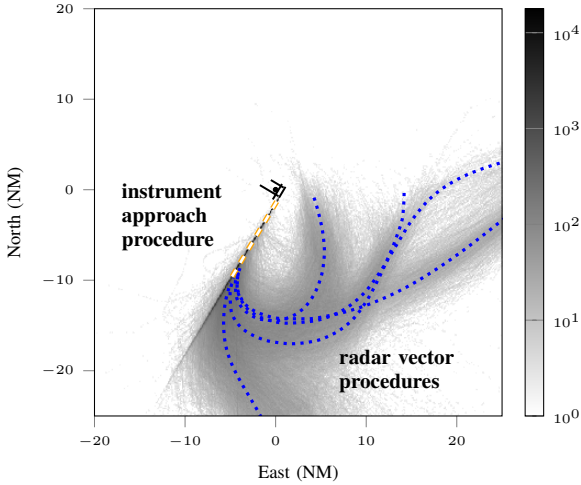}
\caption{Log-histogram of actual arrival tracks to KJFK 04R and associated flight procedures.}
\label{fig:jfk_04R_orginal_paths}
\end{figure}

For KJFK, only the final approach and radar vectoring stages fall in the range of 25 NM from the airport.
Since there are no standard procedures published for radar vectoring, we take the nominal paths extracted from the actual trajectory set as the radar vector procedures.
To extract the nominal paths, we cluster the trajectory set using $k$-means and then select the ones representative of the flight patterns.
Fig. \ref{fig:jfk_04R_orginal_paths} shows the arrival traffic to runway 04R of KJFK and the associated procedures overlaid on them.
The radar vector procedures are indicated with blue dotted lines, while the IAP is indicated with a white dashed line with orange edges.

\section{Single Trajectory Model}
\label{sec:single trajectory model}
This section outlines our method to learn the distribution of aircraft trajectories relative to their flight procedures and to generate synthetic traffic scenarios. 
The interactions between multiple aircraft are not considered in this section. Our approach follows the following steps, which are described in detail in this section. First, we segment our aircraft trajectories to \textbf{construct input vectors} for our Gaussian mixture models. Next, we use these input vectors to \textbf{train a Gaussian Mixture Model} for each flight stage. To avoid overfitting and manage noise within our GMMs, we \textbf{use low-rank approximations of the GMM covariance matrices}. Finally, we use these approximated matrices to \textbf{generate synthetic trajectories}. \Cref{fig:flowchart} illustrates this process. 

\begin{figure}
    \centering
    \tikzstyle{process} = [rectangle, rounded corners, minimum width=3cm, minimum height=1cm, minimum width=5.5cm, text centered, draw=black]
    \tikzstyle{arrow} = [thick, ->, >=stealth]
    \begin{tikzpicture}[node distance=2cm]
    
        \node (step1) [process] {Segment aircraft trajectories};
        \node (step2) [process, below of=step1] {Fit GMMs};
        \node (step3) [process, below of=step2] {Compute low-rank approximation};
        \node (step4) [process, below of=step3] {Generate synthetic trajectories};
    
        \draw [arrow] (step1.south) -- (step2.north) node[midway,right] {Trajectory segments};
        \draw [arrow] (step2.south) -- (step3.north) node[midway,right] {GMMs};
        \draw [arrow] (step3.south) -- (step4.north) node[midway,right] {Low-rank covariance matrices};
    
    \end{tikzpicture}
        \caption{Flowchart overview of single trajectory model\label{fig:flowchart}}
\end{figure}

\subsection{Construction of Input Vector}
\label{subsec:input vector construction}
Before training a separate Gaussian mixture model for each flight stage, we need to construct the input vector sets in the proper format. 
First, each aircraft trajectory is divided into radar vector and final approach segments based on the distance to its IAP.
Then, the radar vector trajectory is assigned to one of the radar vector procedures using dynamic time warping (DTW) \cite{Muller2007}.

DTW is an algorithm for measuring similarity between two temporal sequences, which may vary in length.
It uses a dynamic programming approach to find the shortest distance between these sequences.
Given a pair of vectors ${x_1=[x_1^{(1)},\ldots, x_1^{(m)}] \in \mathbb{R}^m}$ and ${x_2=[x_2^{(1)}, \ldots, x_2^{(n)}] \in \mathbb{R}^n}$, the DTW distance between them is computed as
\begin{equation}
    DTW(x_1, x_2) = D(m, n)
\end{equation}
where for all ${i \in \{1,\ldots,m\},\ j \in \{1,\ldots,n\}}$,
\begin{equation}
    D(i, j) = \lVert x_1^{(i)} - x_2^{(j)} \rVert_2 + \min \begin{cases}
        D(i, j-1) \\ D(i-1, j) \\ D(i-1, j-1).
    \end{cases}
\end{equation}
After measuring the DTW distances between the radar vector trajectory and each of the radar vector procedures, we label the trajectory as the procedure to which it is closest, as measured by DTW distance.

Now with the two sets of aircraft trajectory-procedure pairs, one for each flight stage, we would like to train a GMM that takes as input the sequence of deviations between aircraft positions and the procedure points.
One challenge for training a GMM is that all the input vectors are required to be of the same length.
To deal with the issue of varying lengths of trajectories, we separately interpolate each dimension in an aircraft trajectory as a polynomial function of time and then re-sample a fixed number of points.
We also generate the procedural trajectories by interpolating the waypoints of the procedures and re-sampling the same number of points.
To integrate time into the spatial procedural trajectories of IAPs, we extract aircraft trajectories that pass very close to all the waypoints and take their mean. This allows us to estimate how a trajectory projected on a nominal path would proceed over time. 
The radar vector procedures already involve temporal factors because the procedures are defined as the nominal paths extracted from aircraft trajectories.

\added{As done in prior work \cite{kochenderfer2010airspace}}, we use the piecewise cubic Hermite interpolation method. \added{We select this method because it has a number of desirable characteristics. First, it ensures that the interpolated function is continuous by fitting a cubic polynomial for each piece of the function and requiring first derivative continuity between pieces. Second, it preserves derivative information such as monotonicity; i.e., where the data is monotonic, the interpolated function will be monotonic. Finally, due to its use of low-degree polynomials, it generally avoids oscillation (Runge's phenomenon) that can be common in interpolation methods using high-degree polynomials. As a result of these properties, it is suitable for interpolating trajectory data and in particular procedural trajectories that should not oscillate between sample points.} \deleted{This method fits a cubic polynomial for each piece of the given function and imposes the continuity of the first derivative.
It preserves monotonicity and avoids oscillation in the intervals where the data is monotonic.
This property makes the piecewise cubic Hermite interpolation method appropriate for interpolating trajectory data, especially the procedural trajectories, which should not oscillate between the sample points.}


Finally, we have two sets of training input vectors. An individual input ${\tau \in \mathbb{R}^{3T+2}}$ is defined as
\begin{align}
\begin{split}
\tau = &[t, d, (x_1-x_1^p), (y_1-y_1^p), (z_1-z_1^p),\\
&\ldots,(x_T-x_T^p), (y_T-y_T^p), (z_T-z_T^p)]
\end{split}
\label{eq:single_model_training_data}
\end{align}
where $t$ and $d$ are the transit time and the total distance of the trajectory,
$[x_{1:T}, y_{1:T}, z_{1:T}]$ are the ENU coordinates of the aircraft trajectory at each timestep from 1 to $T$, and $[x_{1:T}^p, y_{1:T}^p, z_{1:T}^p]$ are the ENU coordinates of the corresponding procedural trajectory. 
We keep the transit time and total distance information of each trajectory so that we can later generate synthetic trajectories with reasonable airspeed.

\subsection{Gaussian Mixture Model}
\label{subsec:GMM}

The Gaussian mixture model (GMM) is a probabilistic generative model that assumes the data points are generated from a mixture of Gaussian distributions.
This model is capable of representing the multimodality and the uncertainty of complex data distributions \cite{kochenderfer2015decision}.
Additionally, GMMs are commonly used across broad domains to generate samples that capture representative characteristics of the training data \cite{liu2019improving, chokwitthaya2020applying, li2021gaussian}. 
These advantages align with our objectives of modeling the aircraft behavior and generating realistic trajectories. 

For each flight stage, we construct a GMM with a set of training input vectors defined as (\ref{eq:single_model_training_data}).
If a single vector $\tau$ is sampled from $K$ mixture components, the marginal probability distribution of $\tau$ is 
\begin{align}
    \label{eq:gmm}
    p(\tau) = \sum_{j=1}^K \pi_j \mathcal{N}(\tau \mid \mu_j, \Sigma_j)
\end{align}
where ${\pi_j}$ are mixing coefficients that must satisfy ${ {\textstyle\sum}_{j=1}^{K} \pi_j=1}$ and ${\pi_j \geq 0}$ for all ${j \in \{1,\ldots,K\}}$.
Each Gaussian density ${\mathcal{N}(\tau  \mid \mu_j, \Sigma_j)}$ is called a component of the mixture.
The maximum likelihood estimates of the parameters ${\{\pi_j, \mu_j, \Sigma_j\}}$ for all $j$ given a dataset of the observations are obtained using the expectation-maximization (EM) algorithm \cite{dempster1977maximum}.

\begin{figure}[tb!]
\centering
\setlength\figureheight{4.8cm}
\setlength\figurewidth{6.3cm}
\input{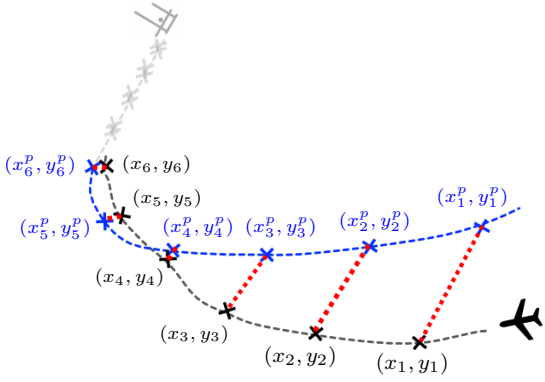}
\caption{Example sequence of deviations between aircraft and procedural trajectory.}
\label{fig:gmm_deviations}
\end{figure}

Our model learns the sequence of deviations (i.e., relative positions) of an aircraft from the corresponding points of the procedure. Fig. \ref{fig:gmm_deviations} shows an example sequence of deviations. 
The aircraft positions and the procedure points are indicated by black and blue crosses respectively, and the deviations are indicated by red dotted lines.

\subsection{Low-rank Approximation of Covariance Matrices}
\label{subsec:low-rank approximation}
Derived from the aircraft and procedural trajectories as in (\ref{eq:single_model_training_data}), the input data matrix of our model is likely to be high-dimensional and contain redundant features.
Also, the model can overfit the noise in the training set. 
To eliminate redundant features and reduce overfitting, we perform a low-rank approximation for each covariance matrix of our GMM in Section \ref{subsec:GMM} using eigenvalue decomposition. 

Consider the covariance of the $j$th Gaussian component ${\Sigma_j \in \mathbb{R}^{n \times n}}$.
The eigenvalue decomposition of ${\Sigma_j}$ is
\begin{align}
    \begin{split}
    \Sigma_j = Q\Lambda Q^{-1} &= Q\Lambda Q^T
    = \sum_{i=1}^n \lambda_i q_i q_i^T
    \end{split}
\end{align}
where ${\Lambda = \text{diag}(\lambda_1, \ldots, \lambda_n) \in \mathbb{R}^{n \times n}}$ is a diagonal matrix with eigenvalues in decreasing order, and ${Q \in \mathbb{R}^{n \times n}}$ is a matrix of the associated eigenvectors.

The best rank-${k}$ (${k \leq n}$) approximation of ${\Sigma_j}$ is obtained by 
\begin{align}
    \widehat{\Sigma}_j = Q_k \Lambda_k Q_k^{-1}
\end{align}
where ${\Lambda_k \in \mathbb{R}^{k \times k}}$ is a diagonal matrix with the largest $k$ eigenvalues, and the columns of ${Q_k \in \mathbb{R}^{n \times k}}$ are the first $k$ eigenvectors.
This is closely related to principal component analysis (PCA) where the $k$ principal axes, a set of orthonormal axes onto which the projection of the original data maximizes variance, are given by the first $k$ eigenvectors.
We can obtain equivalent results from performing a singular value decomposition of $\tau_j$, a set of data vectors assigned to the $j$th Gaussian component.

While the eigenvalue decomposition or PCA provides an analytical solution, both require the rank $k$ to be specified. 
To determine the optimal rank $k^*$ from the observed data rather than setting a specific value for $k$, we adopt the probabilistic principal component analysis (PPCA), a probabilistic formulation of PCA based on a latent variable model \cite{tipping1999mixtures}.

Consider a dataset ${x = \{x_i\}_{i=1}^m \in \mathbb{R}^{m \times n}}$ of $m$ observations.
PPCA assumes that each observation ${x_i \in \mathbb{R}^n}$ is generated from a low-dimensional latent variable ${z_i \in \mathbb{R}^k}$ (${k<n}$) via the following model. For ${i \in \{1,\ldots ,m\}}$, 
\begin{align}
    \label{eq:ppca}
    x_i &= W z_i+\mu+\varepsilon_i
\end{align}
where ${z_i \sim \mathcal{N}(0,I)}$ is a Gaussian latent variable with unit variance, 
${W \in \mathbb{R}^{n \times k}}$ is the weight matrix explaining the dependencies between latent and observed variables,
${\mu \in \mathbb{R}^n}$ is the location parameter that shifts the data,
and ${\varepsilon_i \sim \mathcal{N}(0, \sigma^2 I)}$ is an isotropic Gaussian noise unique to each observed variable.
From (\ref{eq:ppca}), we can compute the following conditional and marginal distributions:
\begin{align}
    \begin{split}
    x_i \mid z_i &\sim \mathcal{N}(W z_i+\mu, \sigma^2 I)\\
    x_i &\sim \mathcal{N}(\mu, WW^T + \sigma^2 I).
    \end{split}
\end{align}

The maximum likelihood estimate (MLE) of the parameters ${\{W, \sigma^2 \}}$ can be solved in closed form \cite{tipping1999mixtures} or using the EM algorithm \cite{roweis1998algorithms}.
The columns of the estimated $W$ define the principal subspace of standard PCA.

The optimal rank $k^*$ can be determined by choosing the latent dimension ${\mathbb{R}^{k}}$ that maximizes the marginal likelihood of the model.


\subsection{Trajectory Generation}
\label{subsec:trajs_generation}
Once we have the GMM parameters for each segment, we can generate synthetic trajectories of aircraft positions based on the trained models and test procedures.
To test the model on a set of procedures not used for the training, we need the procedural trajectories with the relative frequencies of procedures in each segment.

We start generating an aircraft trajectory with its radar vector segment.
First, we randomly select one of the test radar vector procedures with probability proportional to their relative frequencies.
Then, we sample a sequence of deviations ${\tau^v}$ from the radar vector GMM. Next, we construct a trajectory of aircraft positions using the sampled deviations and the test procedural trajectory.
While the total distance an aircraft travels varies with the procedure it follows, the reconstructed trajectory always has the same number of points.
To generate a trajectory with reasonable airspeed, we first compute the adjusted total transit time as ${t'^v = (\tau^v_1 / \tau^v_2) \times d'^v}$
where ${\tau^v_1, \tau^v_2}$ are the transit time and total distance of the sample, and ${d'^v}$ is the total distance of the given test procedure.
Then we align the trajectory with a vector of evenly spaced numbers over the interval ${[0, t'^v]}$.

For a smooth transition between two separately modeled segments of our generated trajectory,
we take the final $n$ positions of the reconstructed radar vector trajectory to compute the first $n$ deviations from the test final approach procedure.
Then, we form a conditional distribution of the final approach GMM to sample the remainder of the final approach segment given the first $n$ measurements.

Suppose the input vector and the Gaussian components in (\ref{eq:gmm}) are partitioned as
\begin{align}
    p\left(
    \begin{bmatrix} 
        \tau_a \\ \tau_b 
    \end{bmatrix}\right) 
    &= \sum_{j=1}^K \pi_j \mathcal{N}\left(
    \begin{bmatrix} 
        \tau_a \\ \tau_b 
    \end{bmatrix} \;\middle|\;
    \begin{bmatrix}
        \mu_{j,a} \\ \mu_{j,b} 
    \end{bmatrix}, 
    \begin{bmatrix}
        \Sigma_{j,aa} \ \Sigma_{j,ab} \\
        \Sigma_{j,ba} \ \Sigma_{j,bb} 
    \end{bmatrix}\right).
\end{align}

Then, the conditional distribution of ${\tau_b}$ given ${\tau_a}$ is
\begin{align}
    \begin{split}  
    p(\tau_b \mid \tau_a) &=
    \sum_{j=1}^K \pi_{j,b \mid a}
    \mathcal{N}\left(\tau_b \mid \mu_{j, b \mid a}, \Sigma_{j, b \mid a}
    \right)\\
    \text{where } \ \pi_{j,b \mid a} &= \frac{\pi_j \mathcal{N}\left( \tau_a \mid \mu_{j,a}, \Sigma_{j,aa} \right)}{\sum_{k=1}^K \pi_k \mathcal{N}\left( \tau_a \mid \mu_{k,a}, \Sigma_{k,aa} \right)} \\
    \mu_{j, b \mid a} &= \mu_{j,b} + \Sigma_{j,ba}\Sigma_{j,aa}^{-1}(\tau_a - \mu_{j,a}) \\
    \Sigma_{j,b \mid a} &= \Sigma_{j,bb} - \Sigma_{j,ba} \Sigma_{j,aa}^{-1} \Sigma_{j,ba}^T.
    \end{split}
    \label{eq:conditional_density_GMM}
\end{align}

\begin{figure}[tb!]
\centering
\input{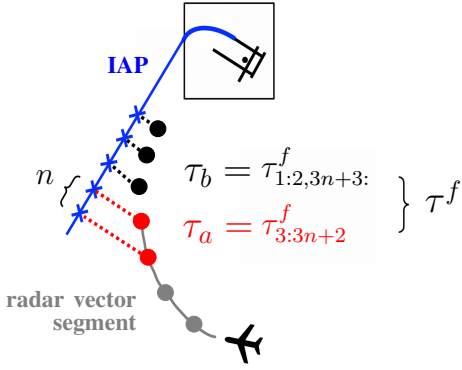}
\caption{Partition of the sequence of deviations for the final approach segment to form a conditional distribution.}
\label{fig:traj_generate_conditional}
\end{figure}

We form this conditional distribution for the final approach GMM by partitioning each vector $\tau^f$ as illustrated in Fig. \ref{fig:traj_generate_conditional}.
In our case, $\tau_a$ is defined as the first $n$ 3D coordinates of deviations and $\tau_b$ corresponds to the remainder of the vector.
In the figure, the blue crosses indicate the procedural points along the IAP. 
The set of dotted lines are the sequence of deviations $\tau^f$ partitioned into $\tau_a$ and $\tau_b$, and
the dots are the reconstructed trajectory of aircraft positions. 
Those marked red correspond to the conditioned part.

After we sample the remaining final approach segment of deviations from the conditional, we reconstruct an aircraft position trajectory as we need for the radar vector segment.
We also repeat the process for integrating time into the trajectory using $\tau^f_1$ and $\tau^f_2$.
Finally, the whole synthetic trajectory is obtained by combining the aircraft trajectories of both segments.

\section{Multiple Trajectory Model}
\label{sec:multiple trajectory model}
When multiple aircraft are within close proximity, their behavior will likely influence each other.
To capture correlations in behavior, we extend the single trajectory GMM described in Section \ref{sec:single trajectory model} to a pairwise trajectory GMM.
Then, we introduce a method to generate multiple trajectories based on the pairwise GMM.
This pairwise approach can scale to large sizes of trajectory set\textcolor{blue}{s} and efficiently generate an arbitrary number of trajectories from a single model, whereas fitting a GMM for the whole set of trajectories may lead to failure in EM due to singularities and quadratic increase in the covariance parameters.
To illustrate, we provide the steps to generate a set of three arrival trajectories.
To define the range of influence, we use the inter-arrival time between each pair of successive arrivals. 

\subsection{Pairwise Trajectory GMM}
For each combination of two procedures, a pairwise trajectory GMM is trained using the trajectory data of two aircraft following their corresponding procedures.
Each training input has the form
\begin{align}
\tau^{pair} = [\tau^{(1)}, \delta^{12}, \tau^{(2)}]
\end{align}
where $\delta^{12}$ is the inter-arrival time between the pair of aircraft, and
\begin{align*}
\tau^{(1)} = \ &[t^1, d^1, (x_1^1-x_1^{1p}), (y_1^1-y_1^{1p}), (z_1^1-z_1^{1p}),\\
& \ \ldots,(x_T^1-x_T^{1p}), (y_T^1-y_T^{1p}), (z_T^1-z_T^{1p})] \in \mathbb{R}^{3T + 2},\\
\tau^{(2)} = \ &[t^2, d^2, (x_1^2-x_1^{2p}), (y_1^2-y_1^{2p}), (z_1^2-z_1^{2p}),\\
& \ \ldots,(x_T^2-x_T^{2p}), (y_T^2-y_T^{2p}), (z_T^2-z_T^{2p})] \in \mathbb{R}^{3T + 2}
\end{align*}
are the sequences of deviations for the first and the second aircraft,
each of which are in the form of
(\ref{eq:single_model_training_data}).

\subsection{Multiple Trajectory Generation}
\label{subsec:construction_of_cov_matrix}


\begin{figure}[tb!]
    \centering
    \includegraphics[width=3.6in]{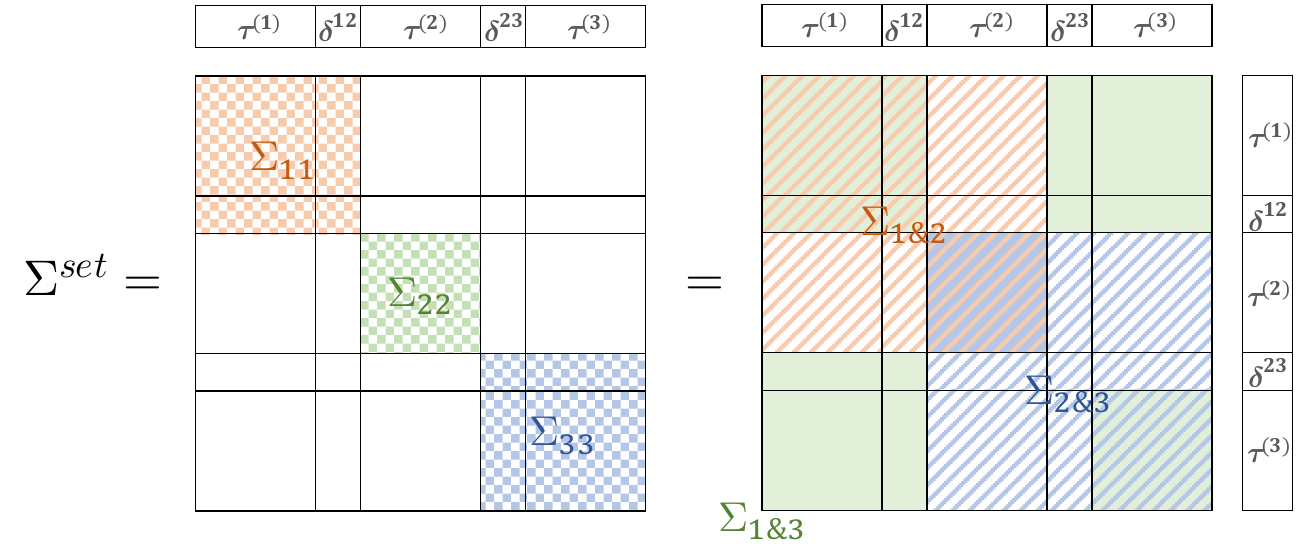}
    \caption{
        Sub-blocks of the covariance matrix $\Sigma_{set}$.
        }
    \label{fig:cov_set_subblocks}
\end{figure}

\begin{figure*}[tb!]
    \centering
    \subfigure[Step 1. Construct $\Sigma_{1\&2}$ by sampling a Gaussian component from the trained pairwise trajectory model.]{
        \label{fig:cov_set_step1}
        \includegraphics[width=2in]{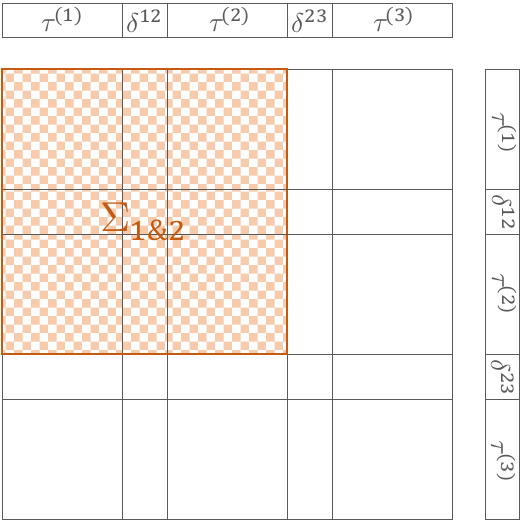}}
    \hspace*{2em}%
    \subfigure[Step 2. Construct $\Sigma_{2\&3}$ by selecting the Gaussian component from the pairwise model that has the closest $\Sigma_{22}$ sub-block from the sampled one in step 1.]{
        \label{fig:cov_set_step2}
        \includegraphics[width=2in]{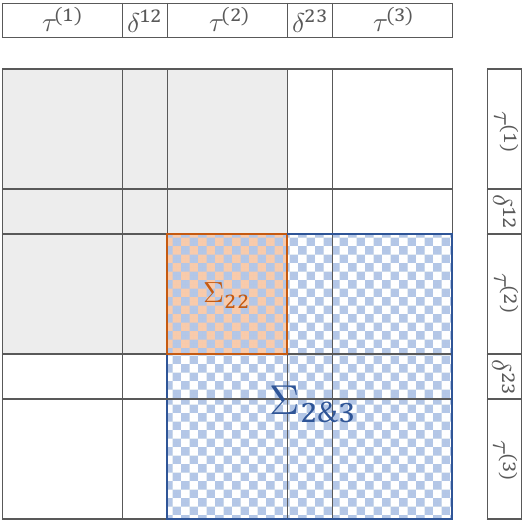}}
    \hspace*{2em}%
    \subfigure[Step 3. Construct $\Sigma_{1\&3}$ by selecting the Gaussian component from the pairwise model that has the closest $\Sigma_{11}$ and $\Sigma_{33}$ sub-blocks from the sampled ones in the previous steps.]{
        \label{fig:cov_set_step3}
        \includegraphics[width=2in]{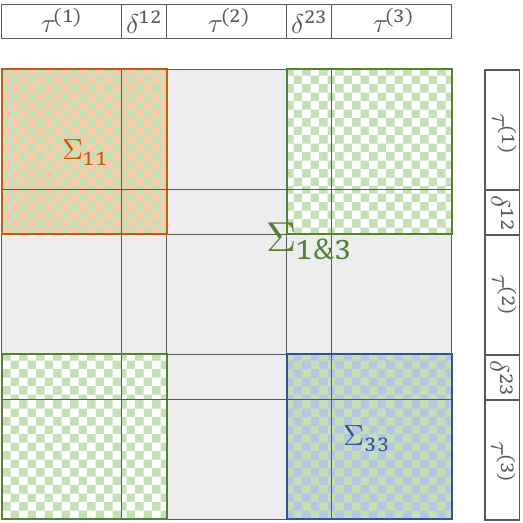}}
    \caption{The process of constructing $\Sigma^{set}$, the covariance matrix of $\tau^{set}$.}
    \label{fig:cov_set_construction}
\end{figure*}

Using the mean vectors and the covariance matrices of the trained pairwise trajectory GMM, we can generate a set of three arrival trajectories that are within the range of influence.
The output vector is
\begin{align}
    \label{eq:input_multiple}
    \tau^{set} = [\tau^{(1)}, \delta^{12}, \tau^{(2)}, \delta^{23}, \tau^{(3)}]
\end{align}
where $\tau^{(1)}$, $\tau^{(2)}$, and $\tau^{(3)}$ are the sequences of deviations for the three aircraft indexed by the order of arrival time, 
and $\delta^{12}$ and $\delta^{23}$ are the inter-arrival times between the aircraft.

We sample $\tau^{set}$ from a Gaussian distribution parameterized by a mean vector $\mu^{set}$ and covariance matrix $\Sigma^{set}$.
To construct $\Sigma^{set}$, we partition the matrix and use its sub-blocks, $\Sigma_{11}, \Sigma_{22}, \Sigma_{33}, \Sigma_{1\&2}, \Sigma_{2\&3}, \text{ and } \Sigma_{1\&3}$. 
These are indicated in Fig. \ref{fig:cov_set_subblocks}.
Then, $\mu^{set}$ and $\Sigma^{set}$ are constructed as follows:
\begin{enumerate}
    \item Construct $\mu^{1\&2}$ and $\Sigma_{1\&2}$ by sampling a Gaussian component from the trained pairwise trajectory model.
    \item Construct $\mu^{2\&3}$ and $\Sigma_{2\&3}$ by selecting the Gaussian component from the pairwise model that has the closest $\Sigma_{22}$ sub-block from the sampled one in step 1.
    \item Construct $\Sigma_{1\&3}$ by selecting the Gaussian component from the pairwise model that has the closest $\Sigma_{11}$ and $\Sigma_{33}$ sub-blocks from the sampled ones in the previous steps.
\end{enumerate}
The corresponding process of constructing $\Sigma^{set}$ is also illustrated in Fig. \ref{fig:cov_set_construction}.

We can scale up to larger number of trajectories by extending this process.
To construct a diagonal pairwise sub-block $\Sigma_{k\&k+1}$, we repeat step 2) to select the Gaussian component that has the closest $\Sigma_{k-1\&k}$ sub-block that is already chosen.
For the other pairwise sub-blocks $\Sigma_{j\&k}$, we repeat step 3) to select the Gaussian component that has the closest $\Sigma_{jj}$ and $\Sigma_{kk}$ sub-blocks that are already constructed in the previous steps.

\section{Experiments}
\label{sec:experiments}
We evaluate the proposed models on the arrival trajectories to KJFK.
We trained the models on the flight tracks and procedures for all the runways except for 13L. Then, the performance of the models \textcolor{blue}{is} validated using 13L data.
This makes an approximately 90--10 training-test split of our data. 
To construct the input vectors as described in Section \ref{subsec:input vector construction}, we set the input length to 150 for the final approach segment and 350 for the radar vector segment.

\subsection{Model Selection}
\label{subsec:model_selection}


The number of clusters is an important hyper-parameter for clustering algorithms including the GMM.
To determine the number of clusters, i.e. Gaussian components in GMMs, we evaluate the cluster performance of our models with different numbers of Gaussian components using the silhouette method.
The silhouette method selects the number of clusters by finding the number that maximizes the silhouette score, i.e. the average of silhouette coefficients over all data points in the entire dataset.
A silhouette coefficient measures how well a data point is matched to its own cluster (cohesion) versus to the other clusters (separation) \cite{rousseeuw1987silhouettes}.
The silhouette coefficient for the $i$th data point is
\begin{align}
s(i) = \frac{b(i)-a(i)}{\max \{ a(i), b(i) \} }
\end{align}
where $a(i)$ is the mean intra-cluster distance (i.e., mean Euclidean distance to the other points in the same cluster) and $b(i)$ is the mean nearest-cluster distance (i.e., mean Euclidean distance to the points in the closest cluster).
The value of $s(i)$ ranges from $-1$ to $+1$, where a higher value indicates that $i$ is well matched to its cluster while poorly to its neighboring cluster. 
If most points have high values, the data points are appropriately clustered.

The silhouette method has several advantages for optimizing the number of clusters. 
It provides an exact solution whereas other methods including the elbow method have to rely on heuristics. 
The silhouette method also quantifies the clustering performance better, as it evaluates both intra-cluster and inter-cluster distances. A good clustering can be characterized by high inter-cluster distance and low intra-cluster distance. \deleted{There are other possible heuristics, which include the Akaike Information Criterion (AIC) and Bayesian Information Criterion (BIC); however, for the aforementioned reasons we use the silhouette method in this work \cite{assess_components2000}.} \added{The silhouette score is illustrated here primarily due to its simplicity compared to heuristic ``elbow'' methods. There are many other possible ways to choose the number of clusters, such as the Akaike Information Criterion (AIC) and the Bayesian Information Criterion (BIC) \cite{assess_components2000}. AIC and BIC tend to give similar results as the silhouette score.}

\begin{figure}[bt!]
\centering
\setlength\figureheight{5.5cm}
\setlength\figurewidth{9cm}
\begin{tikzpicture}

\definecolor{color0}{rgb}{0.392156862745098,0.584313725490196,0.929411764705882}

\begin{axis}[
height=\figureheight,
legend cell align={left},
legend entries={{final approach},{radar vectoring}},
legend style={at={(0.97,0.03)}, anchor=south east, draw=black},
tick align=outside,
tick pos=left,
width=\figurewidth,
x grid style={white!69.01960784313725!black},
xlabel={Number of Gaussian components},
xmin=1.15, xmax=19.85,
y grid style={black},
ylabel={Silhouette scores},
ymin=0.26, ymax=0.57,
ytick={0.30, 0.35, 0.40, 0.45, 0.50, 0.55},
tick label style={font=\tiny},
label style={font=\scriptsize},
legend style={font=\scriptsize}
]
\addlegendimage{mark=square*, line width=1pt, color0}
\addlegendimage{mark=square*, line width=1pt, blue, dashed}
\addplot [line width=1pt, color0, mark=square*, mark size=1, mark options={solid}]
table [row sep=\\]{%
2	0.543160999467294 \\
3	0.447885381746271 \\
4	0.44150655956062 \\
5	0.314570032084669 \\
6	0.332451939072078 \\
7	0.441186079661681 \\
8	0.476373440832541 \\
9	0.490450337400935 \\
10	0.530992943464481 \\
11	0.529126425038986 \\
12	0.53115630163428 \\
13	0.529013952826381 \\
14	0.526929042322887 \\
15	0.523626300441002 \\
16	0.518805112545959 \\
17	0.521322903581658 \\
18	0.515069268838517 \\
19	0.518925910263206 \\
};
\addplot [line width=1pt, blue, dashed, mark=square*, mark size=1, mark options={solid}]
table [row sep=\\]{%
2	0.285397566028243 \\
3	0.323408402354702 \\
4	0.475909060085525 \\
5	0.493234614872401 \\
6	0.493661614016874 \\
7	0.484684414694431 \\
8	0.473991067650981 \\
9	0.465647769098023 \\
10	0.464150055667017 \\
11	0.452642576444028 \\
12	0.442960394770543 \\
13	0.438190935025079 \\
14	0.432229102142355 \\
15	0.422802327630602 \\
16	0.423581095476773 \\
17	0.412475495558834 \\
18	0.399606936627248 \\
19	0.392875310312757 \\
};
\end{axis}

\end{tikzpicture}
\caption{Number of Gaussian components vs. silhouette scores.}
\label{fig:silhouette_iap_vector}
\end{figure}
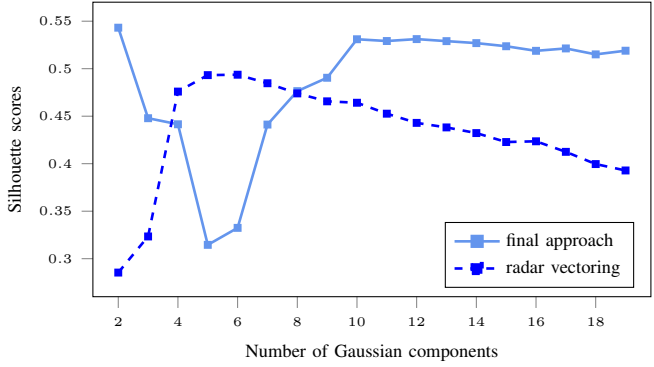

We select the number of Gaussian components that maximize the silhouette score.
Fig. \ref{fig:silhouette_iap_vector} shows the silhouette scores with different numbers of Gaussian components for the final approach model and radar vector model.
The final approach model has the highest score of 0.5432 with two Gaussian components, and the radar vector model has the highest score of 0.4937 with six Gaussian components.

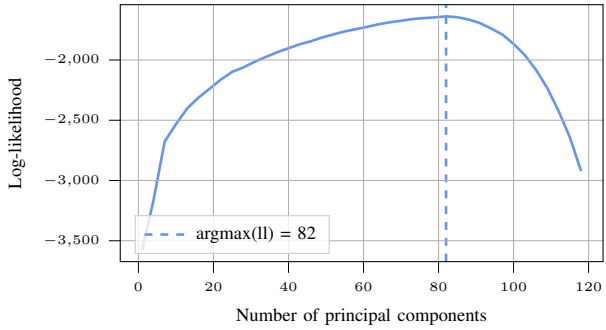
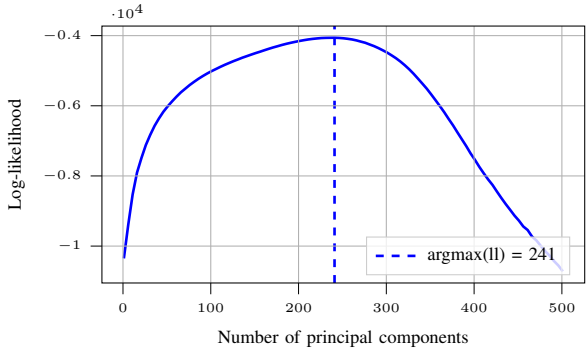
\begin{figure}[bt!]
    \centering
    \subfigure[Final approach model]{
    \setlength\figureheight{5cm}
    \setlength\figurewidth{8cm}
\begin{tikzpicture}

\definecolor{color0}{rgb}{0.392156862745098,0.584313725490196,0.929411764705882}

\begin{axis}[
legend cell align={left},
legend style={fill opacity=0.8, draw opacity=1, text opacity=1, at={(0.03,0.03)}, anchor=south west, draw=white!80!black}, font=\scriptsize,
tick align=outside,
tick pos=left,
x grid style={white!69.0196078431373!black},
xlabel={Number of principal components},
xmajorgrids,
xmin=-4.85, xmax=123.85,
xtick style={color=black},
y grid style={white!69.0196078431373!black},
ylabel={Log-likelihood},
ymajorgrids,
ymin=-3673.16171610547, ymax=-1540.86963896701,
ytick style={color=black},
height=\figureheight,
width=\figurewidth,
tick label style={font=\tiny},
label style={font=\scriptsize}
]
\addplot [line width=1pt, color0, forget plot]
table {%
1 -3576.23934896281
4 -3166.46342875311
7 -2676.37524161896
10 -2532.93526479994
13 -2401.83329424859
16 -2312.24628242538
19 -2237.40096413937
22 -2161.20964576721
25 -2098.32510007955
28 -2061.22182697219
31 -2014.65900535917
34 -1973.7597839886
37 -1935.44639568868
40 -1902.37030692099
43 -1869.55544064166
46 -1844.09827742081
49 -1812.5556854179
52 -1787.23316980128
55 -1762.68212411381
58 -1743.63158220817
61 -1724.7368759245
64 -1703.38068384118
67 -1687.07140172762
70 -1674.45362868706
73 -1660.10416529263
76 -1652.32328136812
79 -1646.26829681805
82 -1637.79200610967
85 -1644.88461469593
88 -1663.75225907883
91 -1693.11227943591
94 -1738.31469100778
97 -1788.36248019015
100 -1866.17047215528
103 -1959.40653891655
106 -2082.63064946482
109 -2233.56695964214
112 -2423.42309932243
115 -2640.96195551629
118 -2921.71241034859
};
\addplot [line width=1pt, color0, dashed]
table {%
82 -3673.16171610547
82 -1540.86963896701
};
\addlegendentry{argmax(ll) = 82}
\end{axis}

\end{tikzpicture}}
    \subfigure[Radar vector model]{
    \setlength\figureheight{5cm}
    \setlength\figurewidth{8cm}
\begin{tikzpicture}

\begin{axis}[
legend cell align={left},
legend style={fill opacity=0.8, draw opacity=1, text opacity=1, at={(0.97,0.03)}, anchor=south east, draw=white!80!black, font=\scriptsize},
tick align=outside,
tick pos=left,
x grid style={white!69.0196078431373!black},
xlabel={Number of principal components},
xmajorgrids,
xmin=-24, xmax=526,
xtick style={color=black},
y grid style={white!69.0196078431373!black},
ylabel={Log-likelihood},
ymajorgrids,
ymin=-11051.7608679236, ymax=-3727.05901271434,
ytick style={color=black},
height=\figureheight,
width=\figurewidth,
tick label style={font=\tiny},
label style={font=\scriptsize}
]
\addplot [line width=1pt, blue, forget plot]
table {%
1 -10351.091372581
6 -9390.29359640526
11 -8527.74752744774
16 -7903.40206105794
21 -7490.5928993267
26 -7120.96882745967
31 -6831.28167651557
36 -6571.9952268888
41 -6359.88297586737
46 -6170.96145818454
51 -6014.11290402957
56 -5872.2950960039
61 -5739.01784809046
66 -5615.47991276329
71 -5511.89210434147
76 -5407.96787815711
81 -5316.5264021807
86 -5229.85766572322
91 -5151.52607292256
96 -5078.44127039378
101 -5011.42107834493
106 -4950.74432311115
111 -4889.44458859846
116 -4831.94946353567
121 -4778.78372448935
126 -4726.38447108067
131 -4678.45004290727
136 -4628.80688123572
141 -4585.33708008752
146 -4541.31427657975
151 -4498.77372454233
156 -4459.26989206967
161 -4417.11993132059
166 -4376.1896841801
171 -4337.37939533848
176 -4302.4885183807
181 -4266.23227598754
186 -4233.9499301315
191 -4204.62757723826
196 -4178.05280795301
201 -4152.24345407912
206 -4129.76973772367
211 -4109.34721272493
216 -4093.00207578259
221 -4080.01033512945
226 -4068.53320608522
231 -4063.447341123
236 -4061.19573093753
241 -4060.00000613294
246 -4065.05278745489
251 -4075.08725256417
256 -4090.41227087023
261 -4109.35525721331
266 -4136.15960424546
271 -4170.77513539739
276 -4203.78637926737
281 -4251.85993686046
286 -4299.111034555
291 -4355.41587096141
296 -4418.19416609366
301 -4484.392697973
306 -4561.92931089684
311 -4643.84898202899
316 -4738.03267096697
321 -4843.10810509342
326 -4957.71811848518
331 -5085.18928061224
336 -5224.23827479698
341 -5365.80307883986
346 -5510.2139115369
351 -5671.6354297764
356 -5838.8209801691
361 -6008.67048258964
366 -6184.53722794659
371 -6375.57497777825
376 -6557.8284285908
381 -6758.27707151676
386 -6953.46289810854
391 -7147.98109913903
396 -7358.01794421731
401 -7543.44595770761
406 -7743.54925260866
411 -7926.39769325349
416 -8100.14000500198
421 -8257.18107727013
426 -8445.39302748818
431 -8627.46085570133
436 -8801.36137819993
441 -8964.28339598023
446 -9130.44110746846
451 -9272.72648759467
456 -9440.08745745432
461 -9542.04406503531
466 -9728.61914326514
471 -9840.02121999606
476 -9993.8857247293
481 -10145.146107708
486 -10273.3291432573
491 -10426.6355094752
496 -10553.6656116963
501 -10718.819874505
};
\addplot [line width=1pt, blue, dashed]
table {%
241 -11051.7608679236
241 -3727.05901271433
};
\addlegendentry{argmax(ll) = 241}
\end{axis}

\end{tikzpicture}}
    \caption{Number of principal components (PC) vs. log-likelihood.}
    \label{fig:low_rank_approximation_iap_vector}
\end{figure}

Once we determine the numbers of Gaussian components, we optimize the ranks for the low-rank approximations of the covariance matrices 
as described in Section \ref{subsec:low-rank approximation}.
Fig. \ref{fig:low_rank_approximation_iap_vector} shows the log-likelihood with different numbers of principal components (PCs) for the final approach model and the radar vector model.
As a result, the 452-dimensional input data of the final approach model can be best represented with 82 PCs, and the 1052-dimensional input data of the radar vector model can be best represented with 241 PCs.

\subsection{Model Validation and Generalization}
\label{subsec:experiment_single_trajectory_model}
\begin{figure}[bt!]
    \centering
    \subfigure[Actual trajectories]{
    \setlength\figureheight{6.7cm}
    \setlength\figurewidth{6.7cm}
    \input{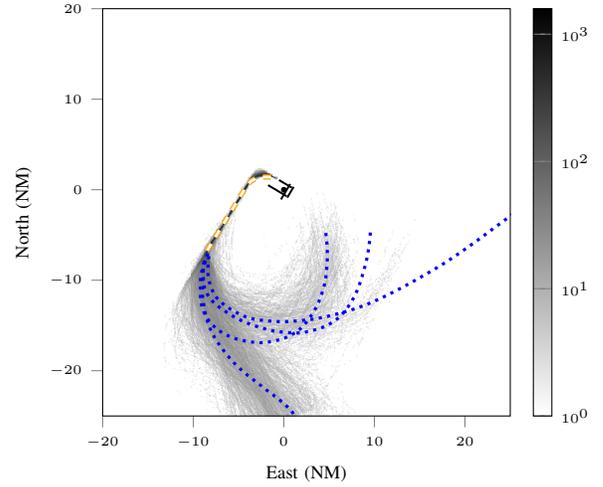}
    \label{fig:jfk_13L_hist_org}}
    \subfigure[Synthetic trajectories]{
    \setlength\figureheight{6.7cm}
    \setlength\figurewidth{6.7cm}
    \input{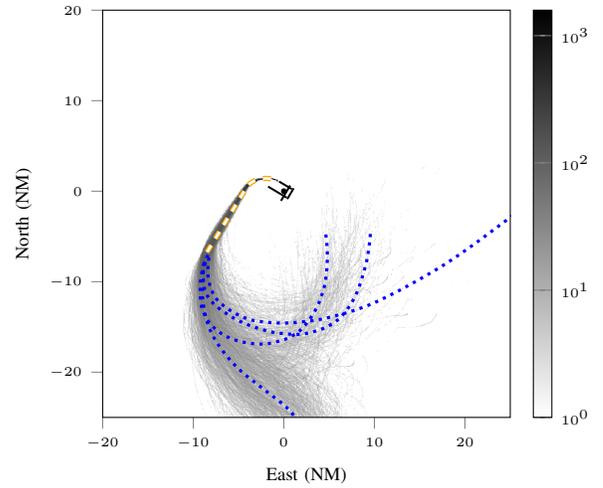}
    \label{fig:jfk_13L_hist_syn}}
    \caption{Log-histograms of arrival tracks to KJFK 13L and associated flight procedures.}
    \label{fig:jfk_13L_org_syn}
\end{figure}

With the optimized parameters, we train the single trajectory model and generate synthetic trajectories from the trained model
as described in Section \ref{sec:single trajectory model}.
Fig. \ref{fig:jfk_13L_hist_org} and Fig. \ref{fig:jfk_13L_hist_syn} show 2D log-histograms of 1,200 actual trajectories and 1,000 synthetic trajectories arriving at KJFK 13L.
As in Fig. \ref{fig:jfk_04R_orginal_paths}, the radar vector procedures and the IAP are indicated in blue dotted lines and a white dashed line with orange edges, respectively.
From the generated trajectories, we observe that the model can learn the general behavior of aircraft with respect to the procedure, but not the distinctive patterns of each individual procedure.
This is because after our model obtains the deviation trajectories with different procedures, it merges all of them into a single dataset for each segment.

\begin{figure}[tb]
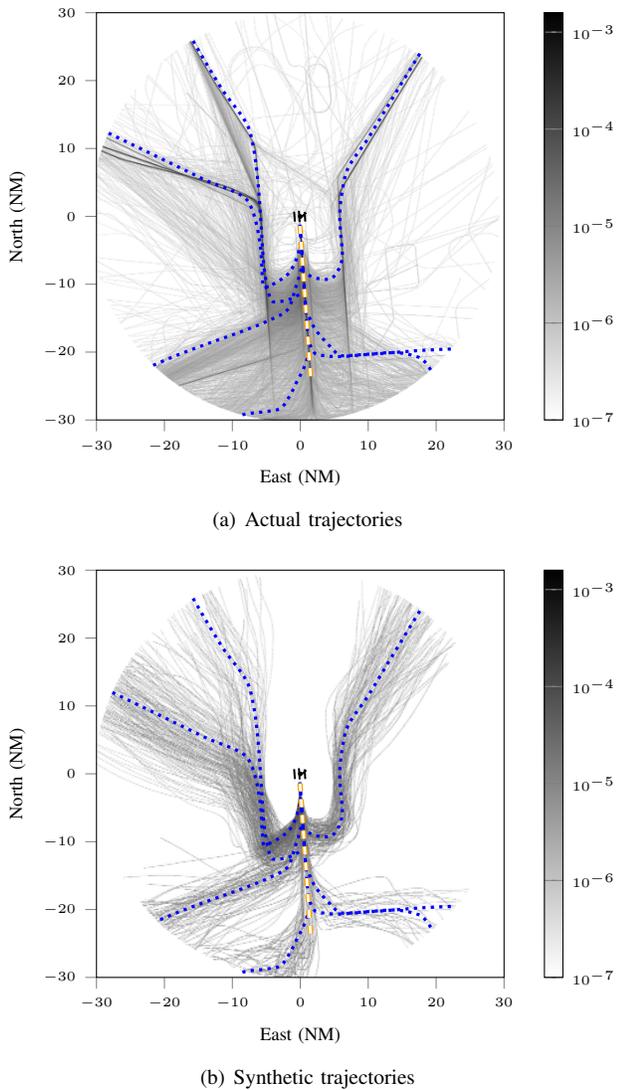

    \centering
    \subfigure[Actual trajectories]{
        \setlength\figureheight{7cm}
        \setlength\figurewidth{7cm}
        \input{figures/clt_36C_original_hist.tex}
        \label{fig:clt_36C_hist_org}}
    \subfigure[Synthetic trajectories]{
        \setlength\figureheight{7cm}
        \setlength\figurewidth{7cm}
        \input{figures/clt_36C_synthetic_hist_500.tex}
        \label{fig:clt_36C_hist_syn}}
  \caption{Log-histograms of arrival tracks to KCLT 36C and RNAV standard arrival procedures.
  }
\end{figure}

To investigate how well the proposed model generalizes to a new environment, we generated synthetic trajectories given Charlotte Douglas Airport (KCLT) runway 36C arrival procedures using the model trained with KJFK data.
Fig. \ref{fig:clt_36C_hist_org} shows a density plot of one month of actual arrivals to KCLT 36C in 2016.
Fig. \ref{fig:clt_36C_hist_syn} is a density plot of 1,000 synthetic trajectories generated by sampling the deviation trajectories from our model and re-constructing the position trajectories with the given flight procedures.
Again, the radar vector procedures and the IAP are indicated in blue dotted lines and a white dashed line with orange edges, respectively.

From the two histograms, we see that some traffic patterns in the actual trajectory set do not appear in the synthetic trajectory set.
Part of the reason is because some trajectories do not appear to follow any of the procedures. 
For example, in Fig. \ref{fig:clt_36C_hist_org}, there is no standard procedure for those trajectories approaching from the northwest, passing north of the airport, and joining the downwind leg on the east side.
Another reason is that the two environments are extremely different.
The radar vector procedures and the actual trajectories of KCLT are mostly straight with only some parts curved, while
those of KJFK do not have straight segments. 
Also, the KJFK trajectories have greater variability in their early arrival stages, while the variability of KCLT trajectories depends on the procedure.
Thus, the synthetic trajectories in Fig. \ref{fig:clt_36C_hist_syn} are more representative of KJFK traffic than KCLT traffic.

We also provide quantitative analyses on the single trajectory model in the next section, along with the multiple trajectory model.

\subsection{Analyses on Multiple Trajectory Scenarios}
To evaluate the performance of the single and multiple trajectory models, we experiment with both models on traffic scenarios where three aircraft arrive at KJFK 13L with inter-arrival times of less than 180 seconds.
As described in Section \ref{sec:multiple trajectory model}, we train a pairwise trajectory model for each combination of two procedures and generate synthetic trajectory sets of three aircraft.
We also generate synthetic trajectory sets using the single trajectory model. 
This is done by generating each trajectory independently and then connecting the trajectory data points through the inter-arrival time variables.

\begin{figure*}[tb!]
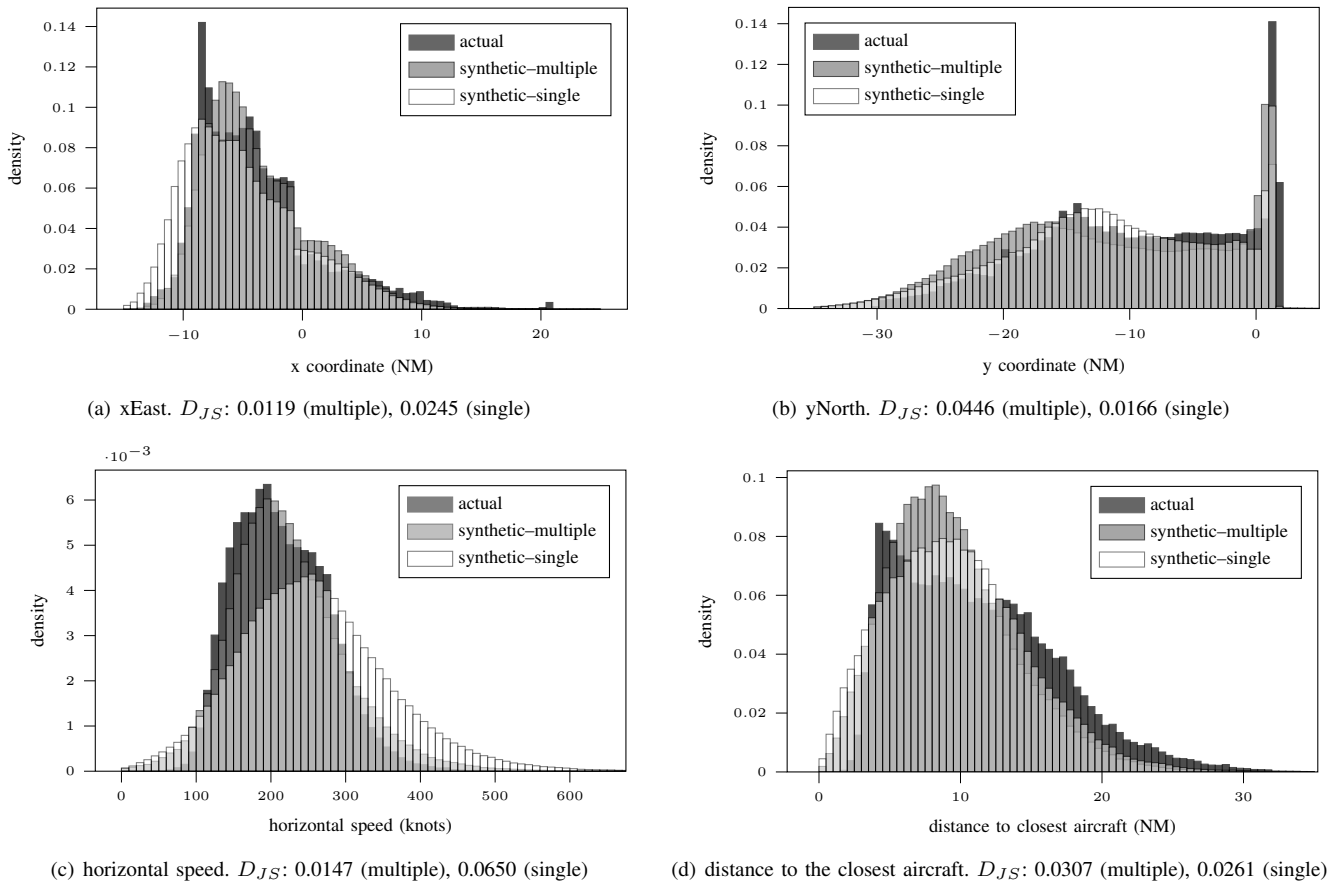

    \centering
    \subfigure[xEast. $D_{JS}$: 0.0119 (multiple), 0.0245 (single)]{
        \label{fig:hist_x_coords}
        \input{figures/hist_x}}
    \hspace*{1em}%
    \subfigure[yNorth. $D_{JS}$: 0.0446 (multiple), 0.0166 (single)]{
        \label{fig:hist_y_coords}
        \input{figures/hist_y}}
    \vspace*{1em}%
    \hspace*{0.8em}%
    \subfigure[horizontal speed. $D_{JS}$: 0.0147 (multiple), 0.0650 (single)]{
        \label{fig:hist_speed}
        \input{figures/hist_hor_speed}}
    \hspace*{1em}%
    \subfigure[distance to the closest aircraft. $D_{JS}$: 0.0307 (multiple), 0.0261 (single)]{
        \label{fig:hist_encounter}
        \input{figures/hist_encounter}}
    \caption{Distributions of position coordinates, horizontal speed, and distance to the closest aircraft (actual vs. synthetic--multiple vs. synthetic--single trajectories of KJFK 13L).}
    \label{fig:result-hist}
\end{figure*}

Fig. \ref{fig:result-hist} shows the empirical distributions of the position coordinates, horizontal speed, and distance to the closest aircraft, for the actual dataset and the synthetic datasets from each model.
From Fig. \ref{fig:hist_speed} and \ref{fig:hist_encounter}, we observe for both variables that the synthetic distribution from the multiple trajectory model (synthetic--multiple) is more concentrated near its mode than the one from the single trajectory model (synthetic--single).
This result is probable because the multiple model not only learns individual trajectories but also the pairwise relations between them. 

We observe from the figures that the synthetic trajectory sets have much higher densities than the actual dataset at lower speeds and ranges.
In practice, most aircraft cannot maintain $100$ knots, which is why the actual dataset has minimal density in this range. 
Also, it is specified in FAA Order JO 7110.65Y \cite{FAAorderATC} that the separation minimum between two IFR aircraft flying below FL$290$ within $40$ miles from the radar antenna is $3$ miles horizontally and $1,000$ feet vertically.
The invalid outputs may occur from the fact that our models do not directly learn the aircraft trajectories, but learn the deviations and then re-construct the trajectories.

To measure the similarity between the actual distribution and each of the synthetic distributions,
we use the Jensen--Shannon divergence $(D_{JS})$, which is a symmetric and smoothed variation of the Kullback--Leibler divergence $(D_{KL})$ \cite{lin1991divergence}.
The $D_{JS}$ between two empirical distributions $P$ and $Q$ is defined as
\begin{align}
    \begin{split}
        D_{JS}(P, Q) &= \frac{1}{2} \Big( D_{KL}(P, M) + D_{KL}(Q, M) \Big)\\
        &= \frac{1}{2}\sum_{x_i} \Big( P(x_i) \log \frac{P(x_i)}{M(x_i)} + Q(x_i) \log \frac{Q(x_i)}{M(x_i)} \Big)
    \end{split}
\end{align}
where ${M = (P+Q)/2}$. The output is bounded by $0$ and $1$.

\begin{table}[ht]
\centering
\caption{$D_{JS}$ between actual distribution, distribution of synthetic trajectories (single model) and distribution of synthetic trajectories (multiple model)\label{tab:djs}}
\begin{tabular}{@{}lrr@{}} \hline
\textbf{Variable} & \textbf{$D_{JS}$ Single} & \textbf{$D_{JS}$ Multiple} \\ \hline
X-coordinate (NM) & 0.0245 &  0.0119 \\ 
Y-coordinate (NM) & 0.0166 & 0.0446 \\ 
Horizontal speed (Knots) &  0.0650 &  0.0147 \\ 
Distance to closest aircraft (NM) & 0.0261 & 0.0307 \\ \hline
\end{tabular}
\end{table}

Table \ref{tab:djs} presents the $D_{JS}$ between the actual distribution and each of the synthetic distributions for different variables (See Fig. \ref{fig:result-hist} for a visualization of the distributions).
We find that the multiple model outperforms the single model in terms of horizontal speed, not only because the multiple model has lower $D_{JS}$ but the single model outputs speeds that are too high or too low.
For the closest distance, the single model has \textcolor{blue}{a }slightly lower JS divergence.
What is more important, however, is the loss of separation that occurs whenever the separation minima are breached.
We counted the number of times when the distance to the closest aircraft goes below $3$ miles within the 1,000 trajectory sets generated using each model.
It turns out that loss of separation occurred $781$ times when using the single model, and $482$ times when using the multiple model.
These results indicate that the multiple model performs better, or at least comparably than the single model in terms of both similarity and safety. We note that an explicit collision avoidance model can be added on top of this nominal trajectory model to prevent actual collisions in simulation.

\begin{figure*}[bt!]
    \centering
    \subfigure[First example set.]{
    \includegraphics[width=7.1in]{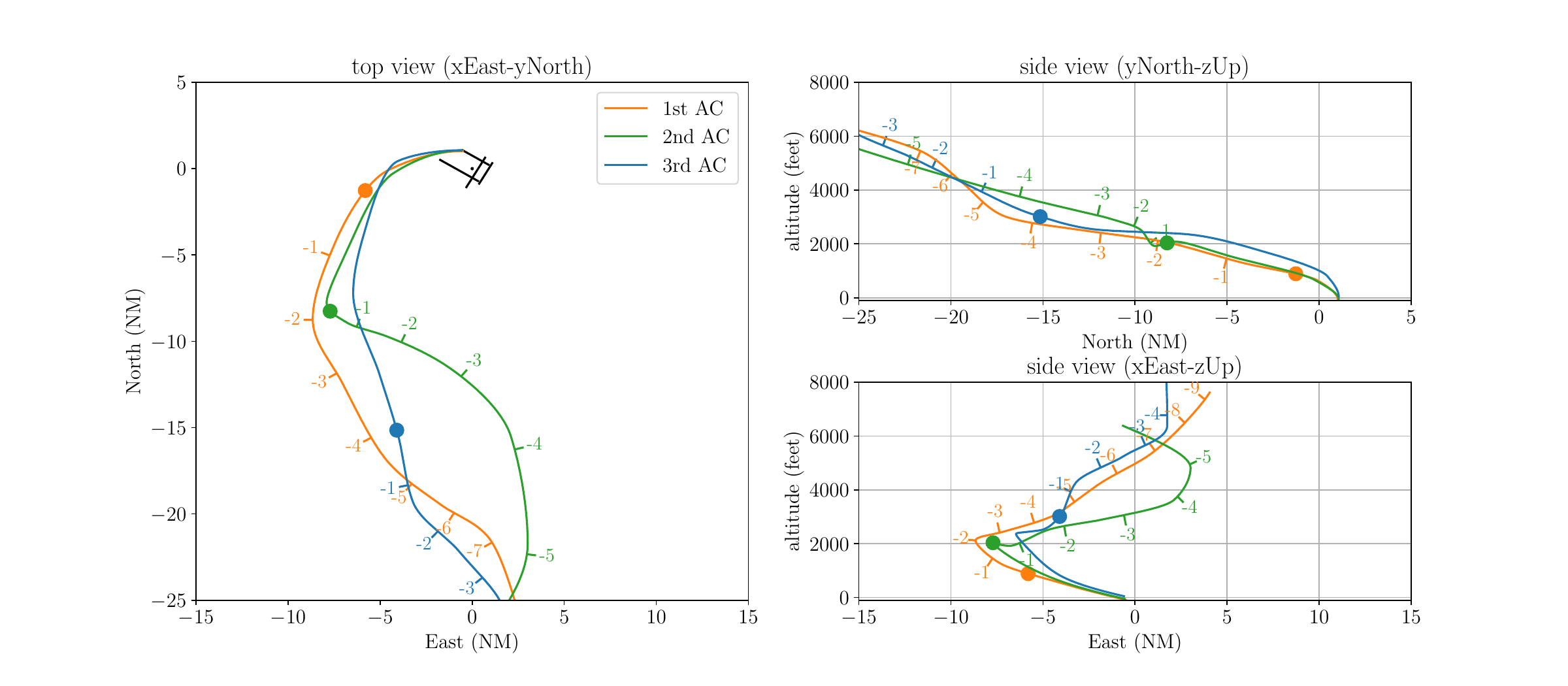}
    \label{fig:multi_trajs_generation_set1}}
    \subfigure[Second example set.]{
    \includegraphics[width=7.1in]{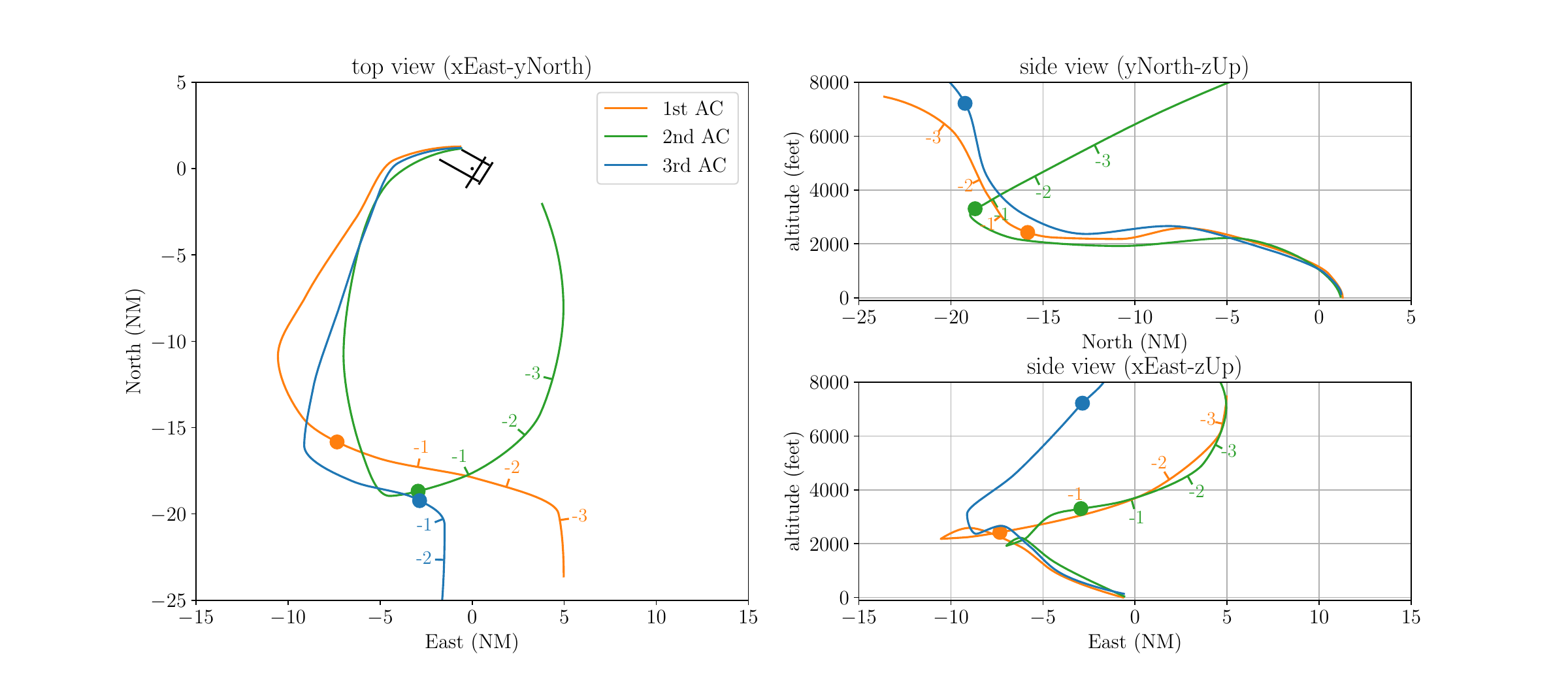}
    \label{fig:multi_trajs_generation_set2}}
    \caption{Example sets of three trajectories generated based on the multiple trajectory model. Top view (left column) and side views from north and east (right column).}
    \label{fig:multi_trajs_generation_sets}
\end{figure*}

Fig. \ref{fig:multi_trajs_generation_sets} shows two sample three-trajectory sets generated using the multiple model\textcolor{blue}{s.}
Along each trajectory, the dots indicate the current positions and the ticks indicate previous positions at each minute.
For example, the ${-1}$ ticks indicate the locations of the three aircraft one minute ago.
For the first set shown in Fig. \ref{fig:multi_trajs_generation_set1}, the model successfully generated the trajectories, satisfying the horizontal and vertical separation requirements.
After the second AC gets close to the first AC, it makes a detour and reduces speed to maintain the horizontal and vertical separation from the first AC.
For the second set shown in Fig. \ref{fig:multi_trajs_generation_set2}, on the other hand, loss of separation occurred between the first AC and the second AC at ${t=-2}$.
Besides, at the current time step, the second AC and the third AC overlap in the top view.
Although the two ACs are maintaining the vertical separation minimum, in practice, the air traffic controllers do not want to see overlapping tracks from their radar control screen.
A plausible reason for this undesirable output is the configuration of the procedures.
As with all the other experiments, each trajectory follows one of the procedures for KFJK 13L shown in Fig. \ref{fig:jfk_13L_org_syn}.
We can see that the procedures themselves overlap with each other.
Thus, the synthetic trajectories for KJFK 13L arrivals are likely to overlap or fail in separation even if our models separate them based on their inter-arrival times. 
We could use some heuristics to solve this problem or experiment on another environment such as KCLT 36C where the procedures do not overlap. For now, we will leave these for future work.

To visualize the behavior of multiple trajectories over time, we developed a dynamic simulation tool embedded with Google Maps.
The source code for the experiments and a simulation demo are publicly available at \\
https://github.com/sisl/terminal\_airspace\_modeling.

\section{Conclusions}
\label{sec:conclusions}

In this paper, we presented a method for modeling aircraft behavior in terminal airspace based on radar flight tracks and procedure data. 
We first partitioned the aircraft trajectories into segments based on the structures of flight procedures.
Then, we investigated Gaussian mixture model (GMM) and conditional Gaussian distributions 
for modeling the deviations of aircraft trajectories from their intended flight procedures.
A low-rank approximation was performed on the covariance matrices in GMMs to remove 
feature redundancy and improve the robustness of our model. 
Furthermore, we proposed an idea of modeling pairwise correlations between aircraft which can efficiently generate traffic scenes involving an arbitrary number of aircraft.
The results showed that the proposed models are able to capture the major patterns as well as uncertainties in aircraft behavior relative to the flight procedures. Furthermore, since the models used are simple, they are more robust, faster to train, and more reliable than their complex counterparts.


There are many potential directions for extending this work. First, while the significance of this work lies in its ability to generate trajectories given any procedural data, the proposed model is not generalizable to flight procedures that have distinctively different characteristics from the ones in the training data. Future study should investigate how the knowledge can be transferred across different traffic environments incorporating minimal procedural information into the model. 
Second, further research could be done to reduce the model complexity without sacrificing performance. Using an autoencoder instead of PPCA would further reduce the feature space dimension as it can encode non-linear relationships between variables. Although they are difficult to interpret, transformer-based architectures also demonstrate promising potential for capturing complex patterns. 
Using distances and relative bearings instead of the ENU coordinates to define the deviations could better represent the aircraft behavior, and thus comprise a more explainable feature set. 
Finally, this work distinguished modeling single trajectories and their pairwise correlations in two separate problems. Future work could evaluate these two models on various scenarios with longer durations and different numbers of aircraft involved to explore how they can be integrated.

\section*{Acknowledgments}
This material is based upon work supported by the Department of the Air Force under Air Force Contract FA8702-15-D-0001.
Opinions, findings, conclusions and recommendations are those of the authors and do not necessarily reflect the views of the Department of the Air Force.
The authors acknowledge the FAA and the MITRE Corporation for sharing the surveillance data.
They thank Evan Maki, Randal Guendel, and Mikhail Krichman from MIT Lincoln Laboratory for their support and assistance.
This article benefited from the work of Shane Barratt, and conversations with Rachael Tompa and Kyle Julian.

\printbibliography

\end{document}